\def\year{2022}\relax
\documentclass[letterpaper]{article} 
\usepackage{aaai22}  
\usepackage{times}  
\usepackage{helvet}  
\usepackage{courier}  
\usepackage[hyphens]{url}  
\usepackage{graphicx} 
\usepackage{natbib}  
\usepackage{caption} 
\DeclareCaptionStyle{ruled}{labelfont=normalfont,labelsep=colon,strut=off} 
\usepackage{algorithm}
\usepackage{algorithmic}

\usepackage{newfloat}
\usepackage{listings}

\usepackage{booktabs}
\usepackage{multirow}
\usepackage{changes}
\usepackage{hyperref}

\floatstyle{ruled}
\newfloat{listing}{tb}{lst}{}
\floatname{listing}{Listing}
\title{Sample Prior Guided Robust Model Learning to Suppress Noisy Labels}
\author{
    Wenkai Chen\textsuperscript{\rm 1}, 
    Chuang Zhu\textsuperscript{\rm 1}, 
    Yi Chen\textsuperscript{\rm 2}, 
    Mengting Li\textsuperscript{\rm 1}, 
    Tiejun Huang \textsuperscript{\rm 3}
}
\affiliations{
    \textsuperscript{\rm 1} Beijing University of Posts and Telecommunications\\
    \textsuperscript{\rm 2} Beijing University Of Technology\\
    \textsuperscript{\rm 3} Peking University \\
    
    wkchen@bupt.edu.cn, czhu@bupt.edu.cn, chenyi@emails.bjut.edu.cn, mtli@bupt.edu.cn, tjhuang@pku.edu.cn
}

\begin{document}
\maketitle
\begin{abstract}
Imperfect labels are ubiquitous in real-world datasets and seriously harm the model performance. Several recent effective methods for handling noisy labels have two key steps: 1) dividing samples into cleanly labeled and wrongly labeled sets by training loss, 2) using semi-supervised methods to generate pseudo-labels for samples in the wrongly labeled set. However, current methods always hurt the informative hard samples due to the similar loss distribution between the hard samples and the noisy ones. In this paper, we proposed PGDF (Prior Guided Denoising Framework), a novel framework to learn a deep model to suppress noise by generating the samples' prior knowledge, which is integrated into both dividing samples step and semi-supervised step. Our framework can save more informative hard clean samples into the cleanly labeled set. Besides, our framework also promotes the quality of pseudo-labels during the semi-supervised step by suppressing the noise in the current pseudo-labels generating scheme. To further enhance the hard samples, we reweight the samples in the cleanly labeled set during training. We evaluated our method using synthetic datasets based on CIFAR-10 and CIFAR-100, as well as on the real-world datasets WebVision and Clothing1M. The results demonstrate substantial improvements over state-of-the-art methods.
\end{abstract}

\section{Introduction}
\label{Introduction}

Deep learning techniques, such as convolutional neural networks (CNNs) and recurrent neural networks (RNNs), have recently achieved great success in object recognition \cite{montserrat2017training}, image classification \cite{krizhevsky2012imagenet}, and natural language processing (NLP) \cite{young2018recent}. Most existing CNN or RNN deep models mainly rely on collecting large scale labeled datasets, such as ImageNet \cite{russakovsky2015imagenet}. However, it is very expensive and difficult to collect a large scale dataset with clean labels \cite{yi2019probabilistic}. Moreover, in the real world, noisy labels are often inevitable in manual annotation \cite{sun2019limited}. Therefore, research on designing robust algorithms with noisy labels is of great significance \cite{wei2019harnessing}.

\begin{figure}[]
  \centering
  \includegraphics[width=3.1in]{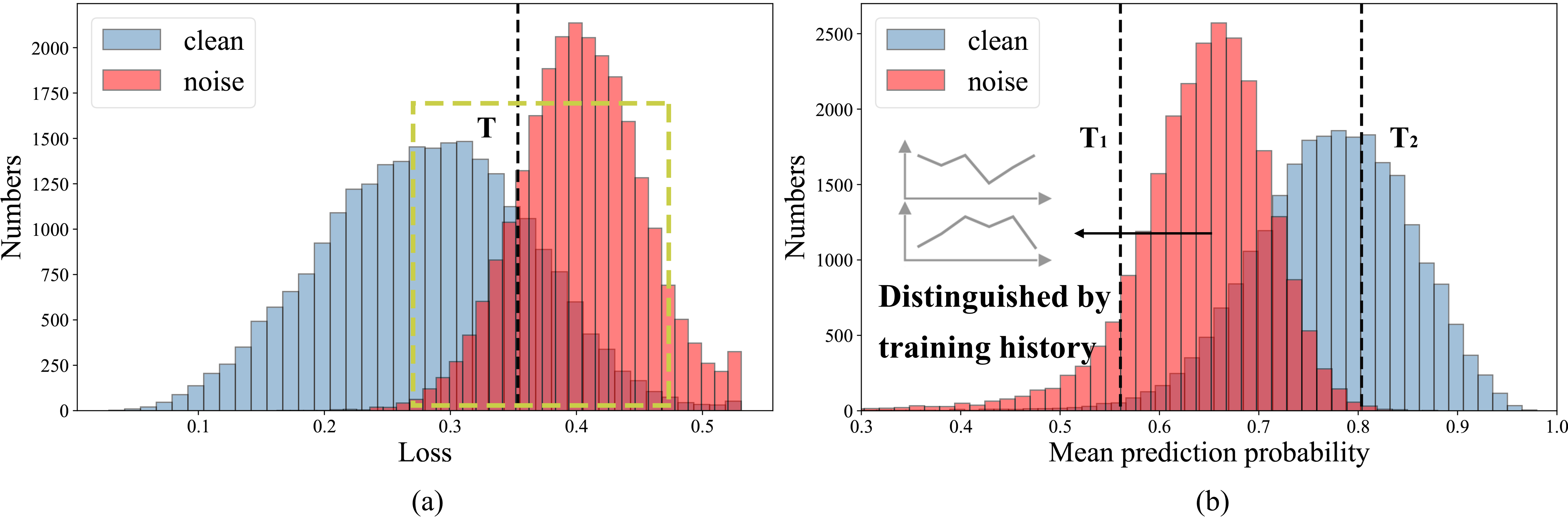}
  \caption{(a) Loss distribution of the clean and noise samples in CIFAR-100 with 50\% symmetric noise ratio. (b) Mean prediction probability distribution of the clean and noisy samples in CIFAR-100 with 50\% symmetric noise ratio.}
  \label{fig:L_MPP}
\end{figure}

In the literature, a lot of approaches were proposed to improve the learning performance with label noise, such as estimating the noise transition matrix \cite{goldberger2016training,patrini2017making}, designing noise-robust loss functions \cite{ghosh2017robust,2019Curriculum,xu2019l_dmi}, designing noise-robust regularization \cite{2020Early,Tanno2020Learning}, sample selection \cite{han2018co,chen2019understanding}, and semi-supervised learning \cite{2020DivideMix,2021LongReMix}. Recently, methods of semi-supervised learning achieve state-of-the-art performance. They always first divide samples into cleanly labeled and wrongly labeled sets by training loss, and then use semi-supervised methods to generate pseudo-labels for samples in the wrongly labeled set. The sample dividing step is generally based on the small-loss strategy \cite{2019How}, where at every epoch, samples with small loss are classified as clean data, and large loss as noise. However, the above methods fail in distinguishing informative hard samples from noisy ones due to their similar loss distributions (as depicted in Figure \ref{fig:L_MPP}(a), samples in the yellow dotted box are indistinguishable), and thus may neglect the important information of the hard samples \cite{xiao2015learning}. To the best of our knowledge, there are very few works studying hard samples under noisy label scenarios. Work \cite{wang2019symmetric} mentioned the hard samples in learning with noisy labels, but that work did not explicitly identify hard samples.

{Although the hard samples and noisy samples cannot be directly distinguished by training loss, we observed that they have different behaviors in training history. Through this intuition, }we propose PGDF (Prior Guided Denoising Framework), a novel framework to learn a deep model to suppress noise. We first use the training history to distinguish the hard samples from the noisy ones (as depicted in Figure \ref{fig:L_MPP}(b), samples between two thresholds are distinguished by training history). Thus, we classify the samples into three sets by using the prior knowledge. The divided dataset is then guiding the subsequent training process. Our key {findings} and contributions are summarized as follows:
\begin{itemize}
    \item Hard samples and noisy samples can be recognized using training history. We first propose a Prior Generation Module, which generates the prior knowledge to pre-classify the samples into an easy set, a hard set, and a noisy set. We further optimize the pre-classification result at each epoch with adaptive sample attribution obtained by Gaussian Mixture Model.
    \item We realize robust noisy labels suppression based on the divided sets. On one hand, we generate high-quality pseudo-labels by the estimated distribution transition matrix with the divided easy set. On the other hand, we further safely enhance the informative samples in the hard set, while previous existing noisy labels processing methods cannot achieve this because they fail to distinguish the hard samples and noisy ones.
    \item We experimentally show that our PGDF significantly advances state-of-the-art results on multiple benchmarks with different types and levels of label noise. We also provide the ablation study to examine the effect of different components.
\end{itemize}

\section{Related Work} 
In this section we describe existing works on learning with noisy labels. Typically, the noisy-label processing algorithms can be classified into five categories by exploring different strategies: estimating the noise transition matrix \cite{goldberger2016training,patrini2017making}, designing noise-robust loss functions \cite{ghosh2017robust,2019Curriculum,xu2019l_dmi,zhou2021asymmetric,zhou2021learning}, adding noise-robust regularization \cite{2020Early,Tanno2020Learning}, selecting sample subset \cite{han2018co,chen2019understanding,huang2019o2u}, and semi-supervised learning \cite{2020DivideMix,2021LongReMix}.

In the first category, different transition matrix estimation methods were proposed in \cite{goldberger2016training,patrini2017making}, such as using additional softmax layer \cite{goldberger2016training}, and two-step estimating scheme \cite{patrini2017making}. However, these transition matrix estimations fail in real-world datasets where the utilized prior assumption is no longer valid \cite{2019Deep}. Being free of transition matrix estimation, the second category targets at designing loss functions that have more noise-tolerant power. Work in \cite{ghosh2017robust} adopted mean absolute error (MAE) which demonstrates more noise-robust ability than cross-entropy loss. The authors in work \cite{xu2019l_dmi} proposed determinant-based mutual information loss which can be applied to any existing classification neural networks regardless of the noise pattern. Recently, work \cite{zhou2021learning} proposed a novel strategy to restrict the model output and thus made any loss robust to noisy labels. Nevertheless, it has been reported that performances with such losses are significantly affected by noisy labels \cite{2018Learning}. Such implementations perform well only in simple cases where learning is easy or the number of classes is small. For designing noise-robust regularization, work in \cite{Tanno2020Learning} assumed the existence of multiple annotators and introduced a regularized EM-based approach to model the label transition probability. In work \cite{2020Early}, a regularization term was proposed to implicitly prevent memorization of the false labels.

Most recent successful sample selection strategies in the fourth category conducted noisy label processing by {selecting clean samples through "small-loss" strategy}. {Work \cite{jiang2018mentornet} pre-trained an extra network, and then used the extra network for selecting clean instances to guide the training.} The authors in work \cite{han2018co} proposed a Co-teaching scheme with two models, where each model selected a certain number of small-loss samples and fed them to its peer model for further training. Based on this scheme, work \cite{chen2019understanding} tried to improve the performance by proposing an Iterative Noisy Cross-Validation (INCV) method. {Work \cite{huang2019o2u} adjusted the hyper-parameters of model to make its status transfer from overfitting to underfitting cyclically, and recorded the history of sample training loss to select clean samples.} This family of methods effectively avoids the risk of false correction by simply excluding unreliable samples. However, they may eliminate numerous useful samples. To solve this shortcoming, the methods of the fifth category based on semi-supervised learning treated the noisy samples as unlabeled samples, and used the outputs of classification models as pseudo-labels for subsequent loss calculations. The authors in \cite{2020DivideMix} proposed DivideMix, which relied on MixMatch \cite{2019MixMatch} to linearly combine training samples classified as clean or noisy. Work \cite{2021LongReMix} designed a two-stage method called LongReMix, which first found the high confidence samples and then used the high confidence samples to update the predicted clean set and trained the model. Recently, work \cite{2021Augmentation} used different data augmentation strategies in different steps to improve the performance of DivideMix, {and work \cite{2022contrast} used a self-supervised pre-training method to improve the performance of DivideMix.}

The above sample selection strategy and semi-supervised learning strategy both select the samples with clean labels for the subsequent training process. All of their selecting strategies are based on the training loss because the clean samples tend to have small loss during training. However, they will hurt the informative hard samples due to the similar loss distribution between the hard samples and the noisy ones. Our work strives to reconcile this gap by distinguishing the hard samples from the noisy ones by introducing a sample prior.

\section{Method}

\begin{figure*}[hbt]
  \centering
  \includegraphics[width=5.0in]{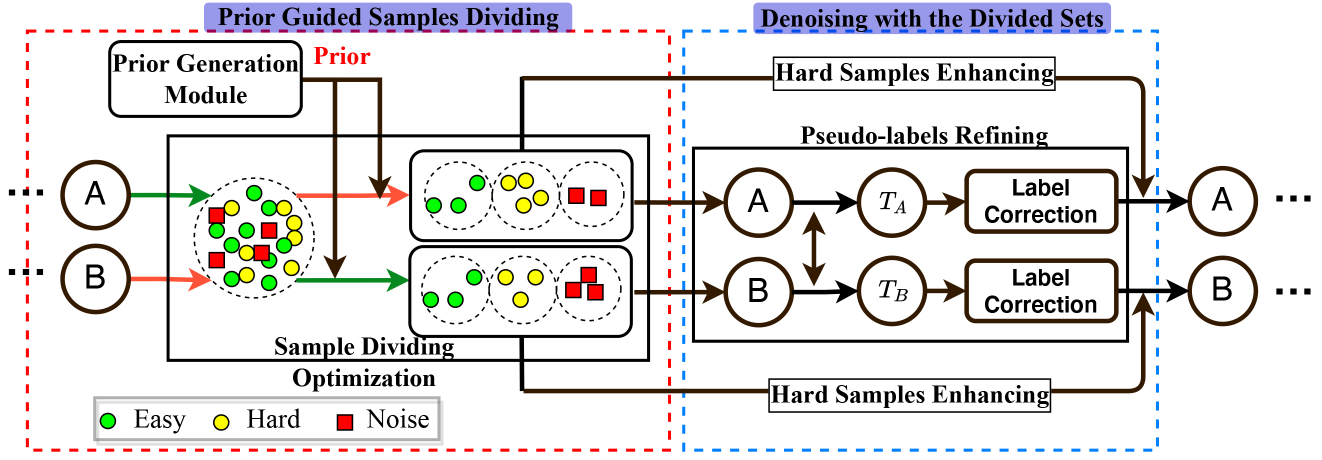}
  \caption{{PGDF first generates the prior knowledge by the Prior Generation Module. Then, it trains two models (A and B) simultaneously. At each epoch, a model divides the original dataset into an easy set, a hard set, and a noisy set by combining the prior knowledge and the loss value of each sample. The divided dataset is used by the other network. After the first stage, the models conduct label correction for samples with the help of the estimated distribution transition matrix. Finally, the training loss is reweighted by the dividing result to further enhance the hard samples.}}
  \label{fig:architecture}
\end{figure*}

\begin{figure*}[hbt]
  \centering
  \includegraphics[width=6.7in]{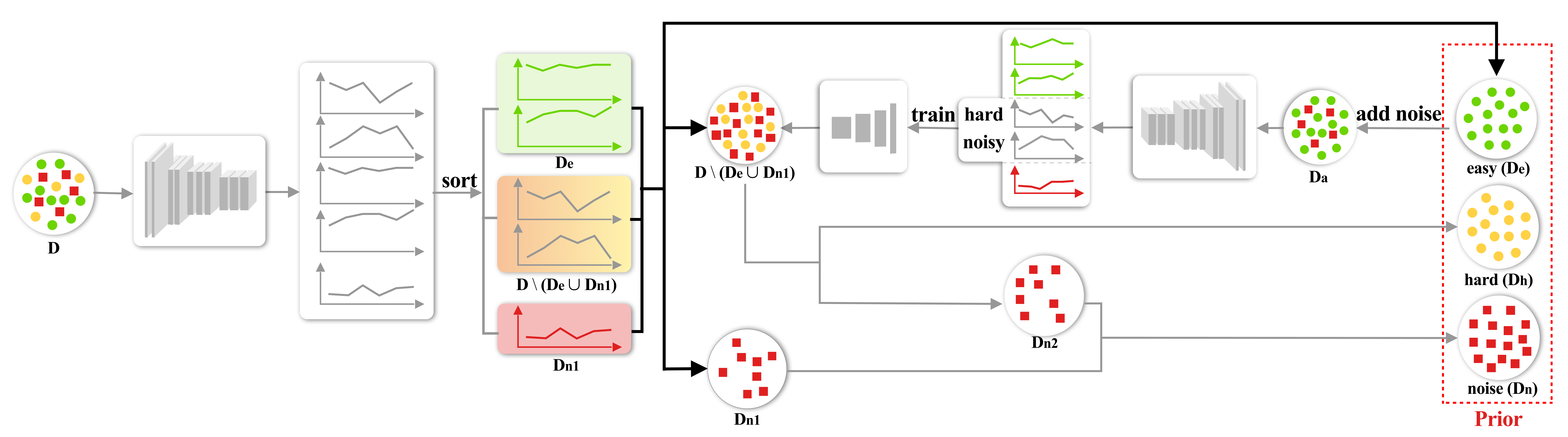}
  \caption{The overview of the Prior Generation Module. It pre-classifies the samples into an easy set, a hard set, and a noisy set.}
  \label{fig:getPrior}
\end{figure*}

{The overview of our proposed PGDF is shown in Figure \ref{fig:architecture}. The first stage (Prior Guided Sample Dividing) of PGDF is dividing the samples into an easy set, a hard set, and a noisy set. The Prior Generation Module first pre-classifies the samples into three sets as prior knowledge; then, at each epoch, the pre-classification result is optimized by the training loss (Sample Dividing Optimization). With the divided sets, the second stage (Denoising with the Divided Sets) conducts label correction for samples with the help of the estimated distribution transition matrix (Pseudo-labels Refining), and then the hard samples are further enhanced (Hard Samples Enhancing). The details of each component are described in the following. }

\subsection{Prior Guided Sample Dividing}
\label{PGM}

\begin{algorithm}
	\renewcommand{\algorithmicrequire}{\textbf{Input:}}
	\renewcommand{\algorithmicensure}{\textbf{Output:}}
	\caption{Prior Generation Module.}
	\label{alg:ehn}
	\begin{algorithmic}[1]
		\REQUIRE  $D=[d_1,d_2,...,d_n]$, $d_i$ is input image, label $Y=[y_1,y_2,...,y_n]$, $y_i$ is label for $d_i$, easy samples ratio $\tau_{e}$, part of noisy samples ratio $\tau_{n1}$
		\ENSURE easy set $D_e$, hard set $D_h$, noise set $D_n$
		\STATE Train classification model $M_c$ by using $D$ and $Y$, record training history $H=[h_1,h_2,...,h_n]$, where $h_i$ is a vector with shape of $1 * k (epoch)$
		\STATE Calculate the mean value of $H$ as $H_m$, $H_m = [mean(h_1),mean(h_2),...,mean(h_n)]$, sort $D$ descending by $H_m$, select easy samples $D_e = D[0:len(D) * \tau_e]$, select part of noisy samples $D_{n1} = D[len(D) * (1-\tau_{n1}):len(D)]$
        \STATE Add noise to $D_e$ as $D_a$, get noisy label $Y_n$ and record whether it is noise or not $R=[r_1,r_2,...,r_n]$
        \STATE Retrain $M_c$ by $D_a$ and $Y_n$, record training history $H_n$
        \STATE Sort $H_n$ descending by mean, select training history $H_n’ = H_n[len(H_n)*\tau_{e}:len(H_n)*(1-\tau_{n1})]$ 
		\STATE Train classifier $M_m$ by using $H_n'$ and $R$ 
        \STATE Put the samples in $D \setminus (D_e \cup D_{n1})$ into $M_m$ and get $D_h$ and $D_{n2}$
        \STATE $D_n = D_{n1} \cup D_{n2}$
		\STATE \textbf{return} $D_e$, $D_h$, $D_n$
	\end{algorithmic}  
\end{algorithm}

\textbf{Prior Generation Module.}
Previous work \cite{2016Understanding} shows that CNNs tend to memorize simple samples first, and then the networks can gradually learn all the remaining samples, even including the noisy samples, due to the high representation capacity. According to this finding, many methods use the ``small-loss'' strategy, where at each epoch, samples with small loss are classified as clean data and large loss as noise. Sample with small loss means the prediction probability of the model output is closer to the supervising label. {We directly used the normalized probability for analysis since the loss value is just calculated by the normalized probability and the ground truth label.} We first train a CNN classification model with the data, and record the probability history of the model output for each sample on the class of its corresponding label. {Then,} we calculate the mean prediction probability value of the sample training history which is shown in Figure \ref{fig:L_MPP}(b). The figure shows the clean sample tends to have a higher mean prediction probability than the noisy one. Therefore, we can set two thresholds (such as the black dotted lines in Figure \ref{fig:L_MPP}(b)). Samples with mean prediction probability lower than $T_1$ are almost noisy, while higher than $T_2$ are almost clean. However, we still cannot distinguish the samples with mean prediction probability between two thresholds. We define this part of clean data as hard samples in our work. 

In order to {distinguish the hard samples from the noisy ones}, we construct the Prior Generation Module based on the prediction history of the training samples, as depicted by Figure \ref{fig:getPrior}. For the training set $D$ with $N$ samples, we gradually obtain the corresponding $N$ prediction probability maps through the training of a CNN classification model for $k$ epochs. This module first selects easy samples $D_e$ and part of noisy samples $D_{n1}$ by using the mean prediction probability values. Then we manually add noise to the $D_e$ as $D_a$ and record whether the sample is noise or not. The noise ratio of the adding noise is the same as the original dataset, {which can be known or estimated by the noise cross-validation algorithm of work \cite{chen2019understanding}.} After that, we train the same classification model by using $D_a$ and record training history again. Then we discard the ``easy samples'' and part of ``noisy samples'' of $D_a$ according to mean prediction probability, and utilize the rest samples as training data to train the classifier. We use a simple one dimension CNN which contains 3 one dimension convolution layers and a fully connected layer as the classifier here. So far, we will obtain a classifier that takes the prediction probability map of training history as input, and output whether it is a hard sample or a noisy one. Finally, we put the samples in $D \setminus (D_e \cup D_{n1})$ into the classifier to get the hard sample set $D_h$ and a part of the noisy set $D_{n2}$, and we combine $D_{n1}$ and $D_{n2}$ as the noisy set $D_n$. Algorithm \ref{alg:ehn} shows the details of this module.


\textbf{Sample Dividing Optimization.} Considering the online training loss at each epoch is also important information to help sample dividing, we apply this information to optimize the pre-classification result. Specifically, as shown in Figure \ref{fig:architecture}, at each epoch, we get the clean probability $w_{it}$ of each sample from training loss by using Gaussian Mixture Model (GMM) following previous work \cite{2020DivideMix}. And we have already got the prior knowledge of each sample, we set the clean probability of prior knowledge as $w_{ip}$ from Equation (\ref{eq:c_p}), 
\begin{equation}
\label{eq:c_p}
w_{ip}=\left\{\begin{array}{cc}
1, & d_{i} \in D_{e} \\
p_{h}, & d_{i} \in D_{h} \\
1-p_{n}, & d_{i} \in D_{n_{2}} \\
0, & d_{i} \in D_{n_{1}} 
\end{array}\right.
,
\end{equation}
where $p_{h}$ is the classifier ($M_m$) prediction probability for $d_{i}$ to be hard sample and $p_{n}$ is the classifier prediction probability for $d_{i}$ to be noisy sample. Then, we combine $w_{it}$ and $w_{ip}$ to get the clean probability $w_i$ by Equation (\ref{eq:wcp}), 
\begin{equation}
\label{eq:wcp}
w_{i}=\left\{\begin{array}{cc}
1, & d_{i} \in D_{e} \\
m w_{i t}+(1-m) w_{i p}, & d_{i} \in D_{h} \cup D_{n}
\end{array}\right.
,
\end{equation}
where $m$ is a hyper-parameter. Finally, we divide samples with $w_i$ equal to $1$ into the easy set $\tilde{D_e}$, the samples with $0.5 < w_i <1$ are divided into the hard set $\tilde{D_h}$, and the rest samples are divided into the noisy set $\tilde{D_n}$. Each network divides the original dataset for the other network to use to avoid confirmation bias of self-training similar to previous works \cite{han2018co,chen2019understanding,2020DivideMix}.

\subsection{Denoising with the Divided Sets}
\label{PGD}
\textbf{Pseudo-labels Refining.} After the sample dividing phase, we combine the outputs of the two models to generate the pseudo-labels $P$ to conduct label correction, similar to ``co-guessing” in DivideMix \cite{2020DivideMix}. Considering the samples in the easy set are highly reliable, we can use this part of data to estimate the distribution difference between pseudo-label and the ground truth, which can then be used to refine the pseudo-labels. Given the ground truth label $Y$, we use a square matrix $T$ to denote the differences between ground truth label distribution $\hat{Y}$ and pseudo-label distribution $\hat{P}$, thus $\hat{P} = \hat{Y}T$ and $\hat{Y} = \hat{P}T^{-1}$. 

Specifically, we use the easy set ${\tilde{D_e}}$ and its label $Y_{\tilde{D_e}}$ to estimate the $T$ to refine $P$. We first pass the easy set ${\tilde{D_e}}$ to the model and get $P_{\tilde{D_e}}$, where $P_{\tilde{D_e}}$ denotes the model outputs of $\tilde{D_e}$. 
Then we obtain $T$, of which the element $T_{i,j}$ can be calculated by Equation (\ref{eq:t_m1}),
\begin{equation}
\label{eq:t_m1}
T_{i, j}=\frac{1}{\left|N_{i}\right|} \sum_{n \in N_{i}} p^n_{j} ,
\end{equation}
where $N_i$ consists of samples with the same label of class $i$ in $D_e$, $\left|N_{i}\right|$ is the sample number of $N_i$, $p^n_{j}$ is the model output softmax probability for class $j$ of the sample $n$. After that, we refine the pseudo-labels $P$ by Equation (\ref{eq:refine_p}), 
\begin{equation}
\label{eq:refine_p}
\tilde{P}= PT^{-1}
,
\end{equation}
where $\tilde{P}$ is the refined pseudo-labels. Because $\tilde{P}$ may contain negative values, we first utilize Equation (\ref{eq:max}) to enable the non-negative matrix, and then perform normalization along the row direction by Equation (\ref{eq:norm}) to ensure the summation of elements in each pseudo-label probability vector equal to $1$.

\begin{equation}
\label{eq:max}
\tilde{P}=\max \left(\tilde{P}, 0\right)
.
\end{equation}

\begin{equation}
\label{eq:norm}
\tilde{P}_{i j}=\tilde{P}_{i j} / \sum_{j} \tilde{P}_{i j}
.
\end{equation}

Finally, the labels of samples in noisy set $\tilde{D_n}$ are replaced by the refined pseudo-labels $\tilde{P}$. And the label of sample $i$ in hard set $\tilde{D_h}$ is replaced by the combination of the refined pseudo-label $p_i$ in $\tilde{P}$ and original label $y_i$ in $Y$ as Equation (\ref{eq:combine}), where $w_i$ is the clean probability of sample $i$.

\begin{equation}
\label{eq:combine}
y_i = w_iy_i + (1-w_i)p_i
.
\end{equation}

\textbf{Hard Sample Enhancing.} After generating the pseudo-labels, the samples in easy set and hard set are grouped in labeled set $\hat{\mathcal{X}}$, and the noisy set is considered as unlabeled set $\hat{\mathcal{U}}$. We followed MixMatch \cite{2019MixMatch} to ``mix'' the data, where each sample is randomly interpolated with another sample to generate mixed input $x$ and label $p$. MixMatch transforms $\hat{\mathcal{X}}$ and $\hat{\mathcal{U}}$ to $\mathcal{X}^{\prime}$ and $\mathcal{U}^{\prime}$. To further enhance the informative hard samples, the loss on $\mathcal{X}^{\prime}$ is reweighted by $w_i$ as shown in Equation (\ref{eq:lx}), where $r$ is a hyper-parameter. Similar to DivideMix \cite{2020DivideMix}, the loss on $\mathcal{U}^{\prime}$ is the mean squared error as shown in Equation (\ref{eq:lu}), and the regularization term is shown in Equation (\ref{eq:lreg}).

\begin{equation}
\label{eq:lx}
\mathcal{L}_{\mathcal{X}}=-\frac{1}{\left|\mathcal{X}^{\prime}\right|} \sum_{x_i \in \mathcal{X}^{\prime}} \frac{1}{w_i^r} \sum_{c} p_{c} \log \left(\mathrm{p}_{\mathrm {model }}^{\mathrm{c}}(x_i ; \theta)\right)
.
\end{equation}

\begin{equation}
\label{eq:lu}
\mathcal{L}_{\mathcal{U}}=\frac{1}{\left|\mathcal{U}^{\prime}\right|} \sum_{x_i \in \mathcal{U}^{\prime}}\left\|p-\mathrm{p}_{\mathrm{model}}(x_i ; \theta)\right\|_{2}^{2}
.
\end{equation}

\begin{equation}
\label{eq:lreg}
\mathcal{L}_{\mathrm{reg}}=\sum_{c} \pi_{c} \log \left(\pi_{c} / \frac{1}{\left|\mathcal{X}^{\prime}\right|+\left|\mathcal{U}^{\prime}\right|} \sum_{x_i \in \mathcal{X}^{\prime}+\mathcal{U}^{\prime}} \mathrm{p}_{\mathrm{model}}^{\mathrm{c}}(x_i ; \theta)\right)
.
\end{equation}

Finally the total loss is defined in Equation (\ref{eq:l}). $\lambda_{u}$ and $\lambda_{r}$ follow the same settings in DivideMix.

\begin{equation}
\label{eq:l}
\mathcal{L}={\mathcal{L}_{\mathcal{X}}}+\lambda_{u} \mathcal{L}_{\mathcal{U}}+\lambda_{r} \mathcal{L}_{\mathrm{reg}}
.
\end{equation}

\section{Experiment}
\subsection{Datasets and Implementation Details}
\label{implementation}

We compare our PGDF with related approaches on four benchmark datasets, namely CIFAR-10 \cite{2009Learning}, CIFAR-100 \cite{2009Learning}, WebVision \cite{2017WebVision}, and Clothing1M \cite{2015Learning}. Both CIFAR-10 and CIFAR-100 have 50000 training and 10000 testing images of size $32\times32$ pixels. And CIFAR-10 contains 10 classes and CIFAR-100 contains 100 classes for classification. As CIFAR-10 and CIFAR-100 datasets originally do not contain label noise, following previous works \cite{2020DivideMix,2021Augmentation}, we experiment with two types of label noise: symmetric and asymmetric. Symmetric noise is generated by randomly replacing the labels for a percentage of the training data with all possible labels, and asymmetric noise is designed to mimic the structure of real-world label noise, where labels are only replaced by similar classes (e.g. deer$\rightarrow$horse, dog$\leftrightarrow$cat) \cite{2020DivideMix}. {WebVision contains 2.4 million images in 1000 classes. Since the dataset is quite large, for quick experiments, we follow the previous works \cite{chen2019understanding,2020DivideMix,wu2021ngc} and only use the first 50 classes of the Google image subset. Its noise level is estimated at 20\% \cite{2019Prestopping}.} Clothing1M is a real-world dataset that consists of 1 million training images acquired from online shopping websites and it is composed of 14 classes. Its noise level is estimated at 38.5\% \cite{2019Prestopping}.

In our experiment, we use the same backbones as previous methods to make our results comparable. For CIFAR-10 and CIFAR-100, we use an 18-layer PreAct ResNet \cite{2016Identity} as the backbone and train it using SGD with a batch size of 128, a momentum of 0.9, a weight decay of 0.0005, and the models are trained for roughly 300 epochs depending on the speed of convergence. The image augmentation strategy is the same as in work \cite{2021Augmentation}. We set the initial learning rate as 0.02, and reduce it by a factor of 10 after 150 epochs. The warm up period is 10 epochs for CIFAR-10 and 30 epochs for CIFAR-100. 

{For WebVision, we use the Inception-ResNet v2 \cite{2017Inception} as the backbone, and train it using SGD with a momentum of 0.9, a learning rate of 0.01, and a batch size of 32. The networks are trained for 80 epochs and the warm up period is 1 epoch.

For Clothing1M, we use a ResNet-50 with pre-trained ImageNet weights. We train the network using SGD for 80 epochs with a momentum of 0.9, a weight decay of 0.001, and a batch size of 32. The initial learning rate is set as 0.002 and reduced by a factor of 10 after 40 epochs. 

The hyper-parameters proposed in this paper are set in the same manner for all datasets. We set $m = 0.5$, $r = 2$, $\tau_e = 0.5*(1-\tau)$, and $\tau_{n1} = 0.5*\tau$ {($\tau$ is the estimated noise ratio)}}.

\begin{table}[]
\caption{Comparison with state-of-the-art methods in test accuracy (\%) on CIFAR-10 with symmetric noise (ranging from 20\% to 90\%) and 40\% asymmetric noise. Results for previous techniques were directly copied from their respective papers.}
\vskip 0.15in
\scriptsize
\centering

\begin{tabular}{@{}llccccc@{}}
\toprule
Noise type                               & \multicolumn{1}{l|}{}     & \multicolumn{4}{c|}{sym.}                                                          & \multicolumn{1}{l}{asym.} \\ \midrule
Method/Noise ratio                       & \multicolumn{1}{l|}{}     & 20\%          & 50\%          & 80\%          & \multicolumn{1}{c|}{90\%}          & 40\%                      \\ \midrule
\multirow{2}{*}{Cross-Entropy}           & \multicolumn{1}{l|}{best} & 86.8          & 79.4          & 62.9          & \multicolumn{1}{c|}{42.7}          & 85.0                      \\
                                         & \multicolumn{1}{l|}{last} & 82.7          & 57.9          & 26.1          & \multicolumn{1}{c|}{16.8}          & 72.3                      \\ \midrule
\multirow{1}{*}{Mixup}             & \multicolumn{1}{l|}{best} & 95.6          & 87.1          & 71.6          & \multicolumn{1}{c|}{52.2}          & -                         \\
\multirow{1}{*}{\cite{2018mixup}}                                         & \multicolumn{1}{l|}{last} & 92.3          & 77.3          & 46.7          & \multicolumn{1}{c|}{43.9}          & -                         \\ \midrule
\multirow{1}{*}{M-correction}      & \multicolumn{1}{l|}{best} & 94.0          & 92.0          & 86.8          & \multicolumn{1}{c|}{69.1}          & 87.4                      \\
\multirow{1}{*}{\cite{2019Unsupervised}}                                  & \multicolumn{1}{l|}{last} & 93.8          & 91.9          & 86.6          & \multicolumn{1}{c|}{68.7}          & 86.3                      \\ \midrule
\multirow{1}{*}{Meta-Learning}     & \multicolumn{1}{l|}{best} & 92.9          & 89.3          & 77.4          & \multicolumn{1}{c|}{58.7}          & 89.2                      \\
\multirow{1}{*}{\cite{2019Learning}}                                      & \multicolumn{1}{l|}{last} & 92.0          & 88.8          & 76.1          & \multicolumn{1}{c|}{58.3}          & 88.6                      \\ \midrule
\multirow{1}{*}{ELR+}              & \multicolumn{1}{l|}{best} & 95.8          & 94.8          & 93.3          & \multicolumn{1}{c|}{78.7}          & 93.0                      \\
\multirow{1}{*}{\cite{2020Early}}                                         & \multicolumn{1}{l|}{last} & -             & -             & -             & \multicolumn{1}{c|}{-}             & -                         \\ \midrule
\multirow{1}{*}{DivideMix}         & \multicolumn{1}{l|}{best} & 96.1          & 94.6          & 93.2          & \multicolumn{1}{c|}{76.0}          & 93.4                      \\
\multirow{1}{*}{\cite{2020DivideMix}}                                         & \multicolumn{1}{l|}{last} & 95.7          & 94.4          & 92.9          & \multicolumn{1}{c|}{75.4}          & 92.1                      \\ \midrule
\multirow{1}{*}{LongReMix}         & \multicolumn{1}{l|}{best} & 96.2          & 95.0          & 93.9          & \multicolumn{1}{c|}{82.0}          & 94.7                      \\
\multirow{1}{*}{\cite{2021LongReMix}}                                         & \multicolumn{1}{l|}{last} & 96.0          & 94.7          & 93.4          & \multicolumn{1}{c|}{81.3}          & 94.3                      \\ \midrule
\multirow{1}{*}{DM-AugDesc-WS-SAW} & \multicolumn{1}{l|}{best} & 96.3          & 95.6          & 93.7          & \multicolumn{1}{c|}{35.3}          & 94.4                      \\
\multirow{1}{*}{\cite{2021Augmentation}}                                         & \multicolumn{1}{l|}{last} & 96.2          & 95.4          & 93.6          & \multicolumn{1}{c|}{10.0}          & 94.1                      \\ \midrule
\multirow{2}{*}{PGDF (ours)}              & \multicolumn{1}{l|}{best} & \textbf{96.7} & \textbf{96.3} & \textbf{94.7} & \multicolumn{1}{c|}{\textbf{84.0}} & \textbf{94.8}             \\
                                         & \multicolumn{1}{l|}{last}                      & \textbf{96.6} & \textbf{96.2} & \textbf{94.6} & \multicolumn{1}{c|}{\textbf{83.1}}                      & \textbf{94.5}             \\ \bottomrule
\end{tabular}
\label{exp:cifar_10}
\vskip -0.1in
\end{table}

\subsection{Comparison with State-of-the-Art Methods}
We compare the performance of PGDF with recent state-of-the-art methods: Mixup \cite{2018mixup}, M-correction \cite{2019Unsupervised}, Meta-Learning \cite{2019Learning}, {NCT} \cite{2020Noisy}, ELR+ \cite{2020Early}, DivideMix \cite{2020DivideMix}, {NGC} \cite{wu2021ngc}, LongReMix \cite{2021LongReMix}, and DM-AugDesc-WS-SAW \footnote{The DM-AugDesc-WS-WAW strategy in work \cite{2021Augmentation} is not considered here because of the unstable reproducibility mentioned in authors' github page. ({https://github.com/KentoNishi/Augmentation-for-LNL/})} \cite{2021Augmentation}.

\begin{table}[]
\caption{Comparison with state-of-the-art methods in test accuracy (\%) on CIFAR-100 with symmetric noise (ranging from 20\% to 90\%). Results for previous techniques were directly copied from their respective papers.}
\vskip 0.15in
\scriptsize
\centering
\begin{tabular}{@{}llcccc@{}}
\toprule
Method/Noise ratio                       & \multicolumn{1}{l|}{}     & 20\%          & 50\%          & 80\%          & 90\%          \\ \midrule
\multirow{2}{*}{Cross-Entropy}           & \multicolumn{1}{l|}{best} & 62.0          & 46.7          & 19.9          & 10.1          \\
                                         & \multicolumn{1}{l|}{last} & 61.8          & 37.3          & 8.8           & 3.5           \\ \midrule
\multirow{1}{*}{Mixup}             & \multicolumn{1}{l|}{best} & 67.8          & 57.3          & 30.8          & 14.6          \\
\multirow{1}{*}{\cite{2018mixup}}                                         & \multicolumn{1}{l|}{last} & 66.0          & 46.6          & 17.6          & 8.1           \\ \midrule
\multirow{1}{*}{M-correction}      & \multicolumn{1}{l|}{best} & 73.9          & 66.1          & 48.2          & 24.3          \\
\multirow{1}{*}{\cite{2019Unsupervised}}                                         & \multicolumn{1}{l|}{last} & 73.4          & 65.4          & 47.6          & 20.5          \\ \midrule
\multirow{1}{*}{Meta-Learning}     & \multicolumn{1}{l|}{best} & 68.5          & 59.2          & 42.4          & 19.5          \\
\multirow{1}{*}{\cite{2019Learning}}                                         & \multicolumn{1}{l|}{last} & 67.7          & 58.0          & 40.1          & 14.3          \\ \midrule
\multirow{1}{*}{ELR+}              & \multicolumn{1}{l|}{best} & 77.6          & 73.6          & 60.8          & 33.4          \\
\multirow{1}{*}{\cite{2020Early}}                                         & \multicolumn{1}{l|}{last} & -             & -             & -             & -             \\ \midrule
\multirow{1}{*}{DivideMix}         & \multicolumn{1}{l|}{best} & 77.3          & 74.6          & 60.2          & 31.5          \\
\multirow{1}{*}{\cite{2020DivideMix}}                                         & \multicolumn{1}{l|}{last} & 76.9          & 74.2          & 59.6          & 31.0          \\ \midrule
\multirow{1}{*}{LongReMix}         & \multicolumn{1}{l|}{best} & 77.8          & 75.6          & 62.9          & 33.8          \\
\multirow{1}{*}{\cite{2021LongReMix}}                                         & \multicolumn{1}{l|}{last} & 77.5          & 75.1          & 62.3          & 33.2          \\ \midrule
\multirow{1}{*}{DM-AugDesc-WS-SAW} & \multicolumn{1}{l|}{best} & 79.6          & 77.6          & 61.8          & 17.3          \\
\multirow{1}{*}{\cite{2021Augmentation}}                                         & \multicolumn{1}{l|}{last} & 79.5          & 77.5          & 61.6          & 15.1          \\ \midrule
\multirow{2}{*}{PGDF (ours)}              & \multicolumn{1}{l|}{best} & \textbf{81.3} & \textbf{78.0} & \textbf{66.7} & \textbf{42.3} \\
                                         & \multicolumn{1}{l|}{last}                      & \textbf{81.2} & \textbf{77.6} & \textbf{65.9} & \textbf{41.7} \\ \bottomrule
\end{tabular}
\label{exp:cifar_100}
\vskip -0.1in
\end{table}

\begin{table}[]
\caption{Comparison with state-of-the-art methods trained on (mini) WebVision dataset in top-1/top-5 accuracy (\%) on the WebVision validation set and the ImageNet ILSVRC12 validation set. Results for previous techniques were directly copied from their respective papers.}
\vskip 0.15in
\scriptsize
\centering
\begin{tabular}{@{}l|cc|cc@{}}
\toprule
\multicolumn{1}{l|}{\multirow{2}{*}{Method}} & \multicolumn{2}{c|}{WebVision}                        & \multicolumn{2}{c}{ILSVRC12}               \\ \cmidrule(l){2-5} 
\multicolumn{1}{l|}{}                        & \multicolumn{1}{c}{top1} & \multicolumn{1}{c|}{top5} & \multicolumn{1}{c}{top1} & top5           \\ \midrule
NCT \cite{2020Noisy}                                         & 75.16                     & 90.77                     & 71.73                     & 91.61          \\
ELR+ \cite{2020Early}                                        & 77.78                     & 91.68                     & 70.29                     & 89.76          \\
DivideMix \cite{2020DivideMix}                                   & 77.32                     & 91.64                     & 75.20                     & 90.84          \\
LongReMix \cite{2021LongReMix}                                   & 78.92                     & 92.32                     & -                         & -              \\
NGC \cite{wu2021ngc}                                         & 79.16                     & 91.84                     & 74.44                     & 91.04          \\ \midrule
\textbf{PGDF (ours)}                         & \textbf{81.47}            & \textbf{94.03}            & \textbf{75.45}            & \textbf{93.11} \\ \bottomrule
\end{tabular}
\label{exp:webvision}
\vskip -0.1in
\end{table}

Table \ref{exp:cifar_10} shows the results on CIFAR-10 with different levels of symmetric label noise ranging from 20\% to 90\% and with 40\% asymmetric noise. Table \ref{exp:cifar_100} shows the results on CIFAR-100 with different levels of symmetric label noise ranging from 20\% to 90\%. {Following the same metrics in previous works \cite{2021Augmentation,2021LongReMix,2020DivideMix,2020Early},} we report both the best test accuracy across all epochs and the averaged test accuracy over the last 10 epochs of training. Our PGDF outperforms the state-of-the-art methods across all noise ratios.

{Table \ref{exp:webvision} compares PGDF with state-of-the-art methods on (mini) WebVision dataset. Our method outperforms all other methods by a large margin. Table \ref{exp:c1m} shows the result on Clothing1M dataset. Our method also achieves state-of-the-art performance. The result shows our method also works in real-world situations.}

\subsection{Ablation Study}
\label{Ablation}

\begin{table}[]
\caption{Comparison with state-of-the-art methods in test accuracy (\%) on the Clothing1M dataset. Results for previous techniques were directly copied from their respective papers.}
\vskip 0.15in
\scriptsize
\centering
\begin{tabular}{@{}lc@{}}
\toprule
\multicolumn{1}{l|}{Method}                  & Test Accuracy        \\ \midrule
\multicolumn{1}{l|}{Cross-Entropy}           & 69.21                \\
\multicolumn{1}{l|}{M-correction \cite{2019Unsupervised}}      & 71.00                \\
\multicolumn{1}{l|}{Meta-Learning \cite{2019Learning} }    & 73.47                \\
\multicolumn{1}{l|}{NCT \cite{2020Noisy} }    & 74.02                \\
\multicolumn{1}{l|}{ELR+ \cite{2020Early}}              & 74.81                \\
\multicolumn{1}{l|}{DivideMix \cite{2020DivideMix}}         & 74.76                \\
\multicolumn{1}{l|}{LongReMix \cite{2021LongReMix}}         & 74.38                \\
\multicolumn{1}{l|}{DM-AugDesc-WS-SAW \cite{2021Augmentation}} & 75.11                \\\midrule
\multicolumn{1}{l|}{PGDF (ours)}              & \textbf{75.19}       \\ \bottomrule
\end{tabular}
\label{exp:c1m}
\vskip -0.1in
\end{table}

\begin{table}[]
\caption{Ablation study results in terms of average test accuracy (\%, 3 runs) with standard deviation on CIFAR-10 with 50\% and 80\% symmetric noise.}
\vskip 0.15in
\scriptsize
\centering
\begin{tabular}{@{}ll|c|c@{}}
\toprule
Method/Noise ratio                                &      & 50\%          & 80\%          \\ \midrule
\multirow{2}{*}{PGDF}                             & best & \textbf{96.26  $\pm$  0.09} & \textbf{94.69  $\pm$  0.46} \\
                                                  & last & \textbf{96.15  $\pm$  0.13} & \textbf{94.55  $\pm$  0.25} \\ \midrule
\multirow{2}{*}{PGDF w/o prior knowledge}         & best & 95.55  $\pm$  0.11          & 93.77  $\pm$  0.19          \\
                                                  & last & 95.12  $\pm$  0.19          & 93.51  $\pm$  0.23          \\ \midrule
\multirow{2}{*}{PGDF w/o two networks}         & best & 95.65  $\pm$  0.35          & 93.84  $\pm$  0.73          \\
                                                  & last & 95.22  $\pm$  0.39          & 93.11  $\pm$  0.80     \\ \midrule
\multirow{2}{*}{PGDF w/o sample dividing optimization} & best & 95.73  $\pm$  0.09          & 93.51  $\pm$  0.28         \\
                                                  & last & 95.50  $\pm$  0.12          & 93.20  $\pm$  0.32          \\ \midrule
\multirow{2}{*}{PGDF w/o pseudo-labels refining} & best & 95.78  $\pm$  0.26          & 94.06  $\pm$  0.68         \\
                                                  & last & 95.35  $\pm$  0.27          & 93.71  $\pm$  0.52          \\ \midrule
\multirow{2}{*}{PGDF with ``co-refinement \& co-guessing''} & best & 95.91  $\pm$  0.22          & 94.30  $\pm$  0.51         \\
                                                  & last & 95.66  $\pm$  0.29          & 93.81  $\pm$  0.62          \\ \midrule
\multirow{2}{*}{PGDF w/o hard samples enhancing}          & best & 96.01  $\pm$  0.19          & 94.39  $\pm$  0.28          \\
                                                  & last & 95.78  $\pm$  0.07          & 94.21  $\pm$  0.23           \\ \bottomrule
\end{tabular}
\label{ablation}
\vskip -0.1in
\end{table}

We study the effect of removing different components to provide insights into what makes our method successful. The result is shown in Table \ref{ablation}.

To study the effect of the prior knowledge, we divide the dataset only by $w_{it}$ and change the easy set threshold to 0.95 because there is no value equal to 1 in $w_{it}$. The result shows the prior knowledge is very effective to save more hard samples and filter more noisy ones. By removing the prior knowledge, the test accuracy decreases by an average of about 0.93\%.

To study the effect of the two networks scheme, we train a single network. All steps are done by itself. By removing the two networks scheme, the test accuracy decreases by an average of about 0.91\%.

{To study the effect of the sample dividing optimization, we divide the dataset only by the prior knowledge $w_{ip}$. The result shows that whether a sample is noisy or not depends not only on the training history, but also on the information of the image itself and the corresponding label. Integrating this information can make judgments more accurate. By removing the sample dividing optimization phase, the test accuracy decreases by an average of about 0.68\%.}

To study the effect of the pseudo-labels refining phase, we use the pseudo-labels without being refined by the estimated transition matrix. By removing the pseudo-labels refining phase, the test accuracy decreases by an average of about 0.69\%. {We also evaluate the pseudo-labels refinement method in DivideMix \cite{2020DivideMix} by replacing our scheme with ``co-refinement" and ``co-guessing". By replacing our pseudo-labels refining phase with ``co-refinement" and ``co-guessing", the test accuracy decreases by an average of about 0.49\%.}

To study the effect of the hard samples enhancing, we remove the hard enhancing component. The decrease in accuracy suggests that by enhancing the informative hard samples, the method yields better performance by an average of 0.30\%.

{Among the prior knowledge, two networks, sample dividing optimization, pseudo-labels refining phase, and hard enhancing, the prior knowledge introduces the maximum performance gain. All components have a certain gain.}

\subsection{Generalization to Instance-dependent Label Noise}

\begin{table}[]
\caption{Comparison with state-of-the-art methods in terms of average test accuracy (\%, 3 runs) on CIFAR-10 with instance-dependent label noise (ranging from 10\% to 40\%).}
\vskip 0.15in
\scriptsize
\centering
\begin{tabular}{@{}l|cccc@{}}
\toprule
Method        & 10\%                & 20\%                & 30\%                & 40\%                \\ \midrule
CE            & 91.25 $\pm$ 0.27          & 86.34 $\pm$ 0.11          & 80.87 $\pm$ 0.05          & 75.68 $\pm$ 0.29          \\
Forward       & 91.06 $\pm$ 0.02          & 86.35 $\pm$ 0.11          & 78.87 $\pm$ 2.66          & 71.12 $\pm$ 0.47          \\
Co-teaching   & 91.22 $\pm$ 0.25          & 87.28 $\pm$ 0.20          & 84.33 $\pm$ 0.17          & 78.72 $\pm$ 0.47          \\
GCE           & 90.97 $\pm$ 0.21          & 86.44 $\pm$ 0.23          & 81.54 $\pm$ 0.15          & 76.71 $\pm$ 0.39          \\
DAC           & 90.94 $\pm$ 0.09          & 86.16 $\pm$ 0.13          & 80.88 $\pm$ 0.46          & 74.80 $\pm$ 0.32          \\
DMI           & 91.26 $\pm$ 0.06          & 86.57 $\pm$ 0.16          & 81.98 $\pm$ 0.57          & 77.81 $\pm$ 0.85          \\
CAL          & 90.55 $\pm$ 0.02          & 87.42 $\pm$ 0.13          & 84.85 $\pm$ 0.07          & 82.18 $\pm$ 0.18          \\
SEAL         & 91.32 $\pm$ 0.14          & 87.79 $\pm$ 0.09          & 85.30 $\pm$ 0.01          & 82.98 $\pm$ 0.05          \\ \midrule
\textbf{PGDF} & \textbf{94.09 $\pm$ 0.27} & \textbf{91.85 $\pm$ 0.09} & \textbf{90.64 $\pm$ 0.50} & \textbf{87.67 $\pm$ 0.32} \\ \bottomrule
\end{tabular}
\label{idn}
\vskip -0.1in
\end{table}

{Note that the instance-dependent label noise is a new challenging synthetic noise type and would introduce many hard confident samples. We conducted additional experiments on this noise type to better illustrate the superiority of our method. In order to make the comparison fair, we followed the same metrics and used the same noisy label files in work \cite{chen2021beyond}. We compared the recent state-of-the-art methods in learning with instance-dependent label noise: CAL \cite{zhu2021second}, SEAL \cite{chen2021beyond}; and some other baselines \cite{patrini2017making,han2018co,zhang2018generalized,thulasidasan2019combating,xu2019l_dmi}. Table \ref{idn} shows the experimental result. Result for ``CAL" was re-implemented based on the public code of work \cite{zhu2021second}. Other results for previous 
methods were directly copied from work \cite{chen2021beyond}.}

{According to the experimental results, our method outperforms all other methods by a large margin. This shows the generalization ability of our method is well since it also works in complex synthetic label noise.}

\subsection{Hyper-parameters Analysis}

{In order to analyze how sensitive PGDF is to the hyper-parameters $\tau_e$ and $\tau_{n1}$, we trained on different $\tau_e$ and $\tau_{n1}$ in CIFAR-10 dataset with 50\% symmetric noise ratio. Specifically, we first adjusted the value of $\tau_e$ with fixed $\tau_{n1}$ = 0.25, and thus obtained the sensitivity of PGDF to $\tau_e$. Then we adjusted the value of $\tau_{n1}$ with fixed $\tau_e$ = 0.25, and thus obtained the sensitivity of PGDF to $\tau_{n1}$. We report both the best test accuracy across all epochs and the averaged test accuracy over the last 10 epochs of training, as shown in Table \ref{exp:parA}. The result shows that the performance is stable when changing $\tau_e$ and $\tau_{n1}$ in a reasonable range. Thus, the performance does not highly rely on the pre-defined settings of $\tau_e$ and $\tau_{n1}$. Although their settings depend on the noise ratio, they are still easy to set due to the insensitivity. In fact, we set hyper-parameter $\tau_e$ to select a part of samples which are highly reliable to train the $M_m$ (the classifier to distinguish the hard and noisy samples). The settings of $\tau_e$ and $\tau_{n1}$ are not critical, they just support the algorithm. Analysis of other hyper-parameters is shown in Appendix.}

\begin{table}[]
\caption{Results in terms of average test accuracy (\%, 3 runs) with standard deviation on different ``$\tau_e$" and ``$\tau_{n1}$" on CIFAR-10 with 50\% symmetric noise ratio.}
\vskip 0.15in
\scriptsize
\centering
\begin{tabular}{@{}l|cccc@{}}
\toprule
$\tau_e$   & 0.2         & 0.25                 & 0.3         & 0.35        \\ \midrule
best & 96.09 $\pm$ 0.04 & {96.26 $\pm$ 0.09} & 96.14 $\pm$ 0.08 & 96.11 $\pm$ 0.06 \\
last & 95.98 $\pm$ 0.06 & {96.15 $\pm$ 0.13} & 95.99 $\pm$ 0.10 & 95.93 $\pm$ 0.09 \\ \midrule
$\tau_{n1}$   & 0.2         & 0.25                 & 0.3         & 0.35        \\ \midrule
best & {96.26 $\pm$ 0.12} & {96.26 $\pm$ 0.09} & 96.25 $\pm$ 0.13 & 96.02 $\pm$ 0.07 \\
last & 96.08 $\pm$ 0.11 & {96.15 $\pm$ 0.13} & 96.11 $\pm$ 0.16 & 95.82 $\pm$ 0.10 \\ \bottomrule
\end{tabular}
\label{exp:parA}
\vskip -0.1in
\end{table}

\begin{figure}[]
  \centering
  \includegraphics[width=3.2in]{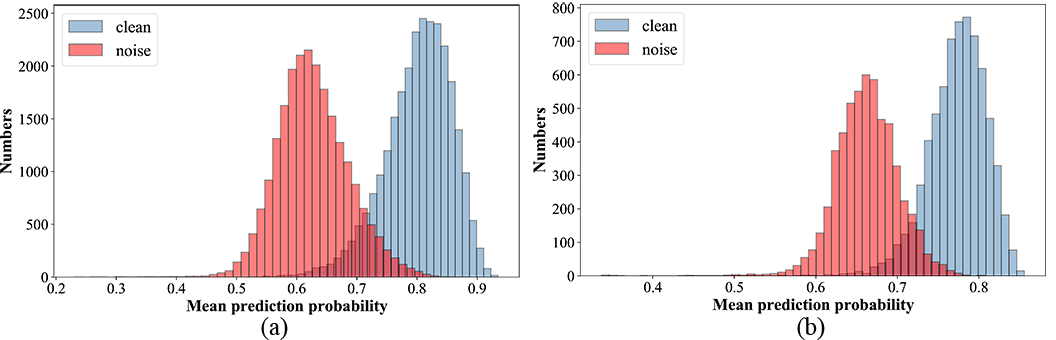}
  \vskip -0.1in
  \caption{{(a) Mean prediction probability histogram of the clean and noisy samples in CIFAR-10 ($D$) with 50\% symmetric noise. (b) Mean prediction probability histogram of the clean and noisy samples in corresponding $D_a$.}}
  \vskip -0.15in
  \label{fig:clean_noise}
\end{figure}

\subsection{Why $D_a$ Works?}
{To study why $M_m$ trained on the artificial created $D_a$ can recognize hard and noisy samples in original dataset $D$, we plotted both the mean prediction probability histogram of the clean and noisy samples in CIFAR-10 (50\% symmetric noise ratio) $D$ and the corresponding $D_a$. The results are shown in Figure \ref{fig:clean_noise}. According to Figure \ref{fig:clean_noise}(b), there are also some clean samples which can not be distinguished from the noisy ones by mean prediction probability. Although the samples in $D_e$ are all easy ones to dataset $D$, part of the samples became hard ones to dataset $D_a$. In Figure \ref{fig:clean_noise}, it should be noted that the mean prediction probabilities of samples trained on $D_a$ have similar distribution with the original dataset $D$, and this is why our Prior Generation Module works by using the artificial created $D_a$.}

\subsection{Work with Other SSL Methods}
{To utilize the samples which were divided into noisy set, we used MixMatch \cite{2019MixMatch} in Section \ref{PGD} while other SSL methods are also applicable. We found that FixMatch \cite{sohn2020fixmatch} as an alternative works even better, and achieves 75.3\% accuracy in Clothing1M dataset. Our method is flexible to other SSL methods. However, these augmentation strategies are not the focus of this paper.}

\section{Limitations}
\label{Limitations}
The quantifiable behavioral differences between hard and noisy samples are not clear. There could exist other better metrics that can be used to directly distinguish them. Our subsequent work will continue to investigate specific quantifiable metrics to simplify the process of the Prior Generation Module, and how to strengthen hard samples more reasonably is a direction worth studying in the future.

\section{Conclusions}
The existing methods for learning with noisy labels fail to distinguish the hard samples from the noisy ones and thus ruin the model performance. In this paper, we propose PGDF to learn a deep model to suppress noise. We found that the training history can be used to distinguish the hard samples and noisy samples. By integrating our Prior Generation, more hard clean samples can be saved. Besides, our pseudo-labels refining and hard enhancing phase further boost the performance. Through extensive experiments show that PGDF outperforms state-of-the-art performance. 


\bibliography{main}

\newpage
\appendix
\onecolumn
\section{Appendix}
\subsection{Additional Experiment Results}
Note that there is another criterion for symmetric label noise injection where the true labels cannot be maintained. Work \cite{2020A} reported the results with the average test accuracy and standard deviation  in this criterion. We use the same backbone (ResNet-18 \cite{2016Deep}) and follow the same metrics as work \cite{2020A}. The results are reported in Table \ref{table:sup}. Our method outperforms all other baselines by a large margin.

\begin{table}[H]
\scriptsize
\centering
\begin{tabular}{@{}l|cccc@{}}
\toprule
Method/Noise ratio & 20\%                                   & 40\%                                   & 60\%                                   & 80\%                                   \\ \midrule
Standard           & 85.7$\pm$0.5                              & 81.8$\pm$0.6                              & 73.7$\pm$1.1                              & 42.0$\pm$2.8                              \\ \midrule
Forgetting(2017)         & 86.0$\pm$0.8                              & 82.1$\pm$0.7                              & 75.5$\pm$0.7                              & 41.3$\pm$3.3                              \\ \midrule
Bootstrap(2014)          & 86.4$\pm$0.6                              & 82.5$\pm$0.1                              & 75.2$\pm$0.8                              & 42.1$\pm$3.3                              \\ \midrule
Forward(2017)            & 85.7$\pm$0.4                              & 81.0$\pm$0.4                              & 73.3$\pm$1.1                              & 31.6$\pm$4.0                              \\ \midrule
Decoupling(2017)         & 87.4$\pm$0.3                              & 83.3$\pm$0.4                              & 73.8$\pm$1.0                              & 36.0$\pm$3.2                              \\ \midrule
MentorNet(2018)          & 88.1$\pm$0.3                              & 81.4$\pm$0.5                              & 70.4$\pm$1.1                              & 31.3$\pm$2.9                              \\ \midrule
Co-teaching(2018)        & 89.2$\pm$0.3                              & 86.4$\pm$0.4                              & 79.0$\pm$0.2                              & 22.9$\pm$3.5                              \\ \midrule
Co-teaching+(2019)       & 89.8$\pm$0.2                              & 86.1$\pm$0.2                              & 74.0$\pm$0.2                              & 17.9$\pm$1.1                              \\ \midrule
IterNLD(2018)            & 87.9$\pm$0.4                              & 83.7$\pm$0.4                              & 74.1$\pm$0.5                              & 38.0$\pm$1.9                              \\ \midrule
RoG(2019)                & {89.2$\pm$0.3}          & {83.5$\pm$0.4}          & {77.9$\pm$0.6}          & {29.1$\pm$1.8}          \\ \midrule
PENCIL(2019)             & {88.2$\pm$0.2}          & {86.6$\pm$0.3}          & {74.3$\pm$0.6}          & {45.3$\pm$1.4}          \\ \midrule
GCE(2018)                & {88.7$\pm$0.3}          & {84.7$\pm$0.4}          & {76.1$\pm$0.3}          & {41.7$\pm$1.0}          \\ \midrule
SL(2019)                 & {89.2$\pm$0.5}          & {85.3$\pm$0.7}          & {78.0$\pm$0.3}          & {44.4$\pm$1.1}          \\ \midrule
TopoCC(2020)             & {89.6$\pm$0.3}          & {86.0$\pm$0.5}          & {78.7$\pm$0.5}          & {43.0$\pm$2.0}          \\ \midrule
TopoFilter(2020)         & {90.2$\pm$0.2}          & {87.2$\pm$0.4}          & {80.5+0.4}           & {45.7$\pm$1.0}          \\ \midrule
PGDF(ours)         & {\textbf{96.50$\pm$0.13}} & {\textbf{96.18$\pm$0.17}} & {\textbf{95.19$\pm$0.38}} & {\textbf{82.15$\pm$1.94}} \\ \bottomrule
\end{tabular}
\caption{Average test accuracy (\%) and stardard deviation on CIFAR-10 with symmetric noise (ranging from 20\% to 80\%). Results for previous techniques were directly copied from \cite{2020A}.}
\label{table:sup}
\end{table}

\subsection{Hyper-parameters Analysis}
In order to analyze the sensitivity of the hyper-parameters proposed in this paper, we conduct experiments on CIFAR-10 dataset with 50\% symmetric noise ratio. Specifically, we first fixed the hyper-parameters settings as $m = 0.5$, $r = 2$, $k = 300$, $\tau_e = 0.25$, and $\tau_{n1} = 0.25$,  then change the value of each hyper-parameter respectively to obtain the sensitivity. The results are shown in Table \ref{exp:parA}. According to Table \ref{exp:parA}, for $m$ and $r$, the performances of the model increase first, then achieve the peak, and finally decrease with the increase of each parameter. For $k$, the performance first increases, and then becomes relatively stable when $k$ exceeds a certain threshold. It should be noted that $m = 0$ is equivalent to the ablation study ``PGDF w/o prior knowledge", and $m = 1$ is equivalent to the ablation study ``PGDF w/o sample dividing optimization".

\begin{table}[H]
\caption{Results in terms of average test accuracy (\%, 3 runs) with standard deviation on different ``$m$", ``$r$" and ``$k$" on CIFAR-10 with 50\% symmetric noise ratio.}
\vskip 0.15in
\scriptsize
\centering
\begin{tabular}{@{}l|ccccc@{}}
\toprule
$m$    & 0           & 0.25                 & 0.5                  & 0.75        & 1           \\ \midrule
best & 95.73$\pm$0.09 & 95.86$\pm$0.14          & {96.26$\pm$0.09} & 95.99$\pm$0.11 & 95.55$\pm$0.11 \\
last & 95.50$\pm$0.12 & 95.73$\pm$0.16          & {96.15$\pm$0.13} & 95.92$\pm$0.12 & 95.13$\pm$0.19 \\ \midrule
$r$    & 1           & 2                    & 3                    & 4           &             \\ \midrule
best & 96.17$\pm$0.07 & {96.26$\pm$0.09} & 96.03$\pm$0.17          & 95.61$\pm$0.23 &             \\
last & 96.00$\pm$0.13 & {96.15$\pm$0.13} & 95.80$\pm$0.23          & 95.40$\pm$0.31 &             \\ \midrule
$k$    & 100         & 200                  & 300                  & 400         &             \\ \midrule
best & 95.74$\pm$0.13 & 96.08$\pm$0.08          & {96.26$\pm$0.09} & 96.23$\pm$0.10 &             \\
last & 95.59$\pm$0.15 & 95.89$\pm$0.13          & {96.15$\pm$0.13} & 96.07$\pm$0.18 &             \\ \bottomrule
\end{tabular}
\label{exp:parA}
\vskip -0.1in
\end{table}

\subsection{Hard and Noisy Sample Behavior Analysis}
\begin{figure}[hbt]
  \centering
  \includegraphics[width=5.0in]{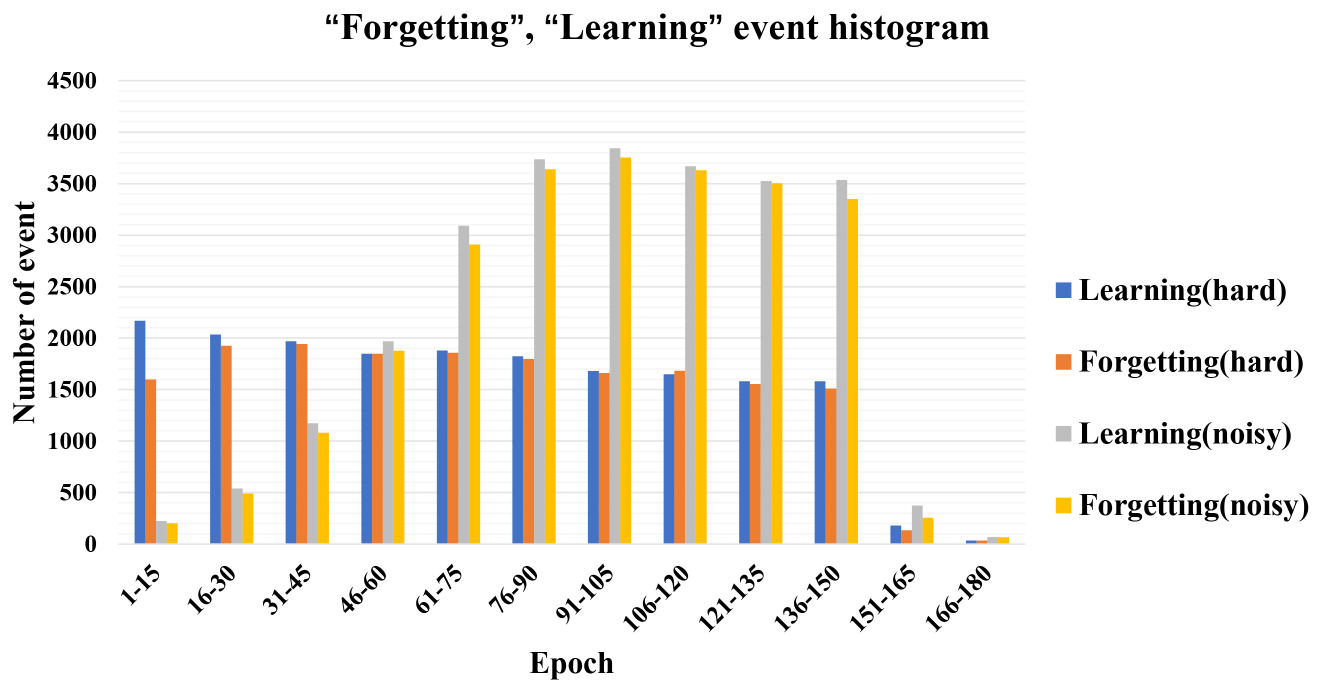}
  \caption{The forgetting and learning event histogram on CIFAR-10 with 20\% symmetric noise.} 
  \label{fig:hsa}
\end{figure}

To analyze the training process behavior of the hard sample and the noise sample, inspired by \cite{toneva2018empirical}, we randomly selected 1000 hard samples, 1000 noisy samples on CIFAR-10 with 20\% symmetric noise, and calculated the number of ``Learning'' event and ``Forgetting'' event of them through the training epochs. The ``Learning'' event in the $t$ epoch is defined as the prediction probability of the labeled class is less than $0.5$ in $t-1$ epoch, while greater than $0.5$ in $t$ epoch. The ``Forgetting'' event in the $t$ epoch is defined as the prediction probability of the labeled class is greater than $0.5$ in $t-1$ epoch, while less than $0.5$ in $t$ epoch. The statistical result is shown in Figure \ref{fig:hsa}. 

This result shows that the hard sample and the noisy sample have great behavioral differences with the increase of the epoch in training. Before the model converges, the number of learning events and forgetting events of noisy samples is relatively small in the early training stage, and gradually increases as the number of training epochs increases. The learning forgetting events of hard samples are relatively consistent throughout the training process.

\subsection{Training Time Report}
We recorded the training time of different steps of PGDF on CIFAR-10 with 50\% symmetric noise by using a single Nvidia V100 GPU. The results are shown in Table \ref{time}.

\begin{table}[H]
\caption{Different steps of training time (hours) on CIFAR-10 with 50\% symmetric noise.}
\vskip 0.15in
\scriptsize
\centering
\begin{tabular}{@{}cc|c@{}}
\toprule
Prior Generation Module & Subsequent training process & All   \\ \midrule
2.1 h                    & 8.2 h                        & 10.3 h \\ \bottomrule
\end{tabular}
\label{time}
\vskip -0.1in
\end{table}

\end{document}